\RequirePackage{amsmath}
\documentclass[runningheads]{llncs}
\usepackage[T1]{fontenc}
\usepackage{graphicx}
\usepackage{amsfonts}
\usepackage{booktabs}
\usepackage{multirow}
\usepackage{subcaption}  
\usepackage{hyperref}

\usepackage{xspace}
\newcommand{\gurobi}{\textit{Gurobi}}
\newcommand{\scip}{\textit{SCIP}}
\newcommand{\smac}{\textit{SMAC}}
\newcommand{\singletask}{\textit{Single-task}\xspace}
\newcommand{\multitaskbapas}{\textit{Multi-task-BAPAS}\xspace}
\newcommand{\multitaskbaconfig}{\textit{Multi-task-BACONFIG}\xspace}
\newcommand{\multitaskpasconfig}{\textit{Multi-task-PASCONFIG}\xspace}

\newcommand{\backdoor}{\textsc{Backdoor}\xspace}
\newcommand{\pas}{\textsc{PaS}\xspace}
\newcommand{\configuration}{\textsc{Configuration}\xspace}

\begin{document}
\title{Multi-task Representation Learning\\ for Mixed Integer Linear Programming}
%
%
\author{Junyang Cai\orcidID{0009-0008-5510-2158}, Taoan Huang\orcidID{0000-0002-8733-2379}, \\Bistra Dilkina\orcidID{0000-0002-6784-473X}}
\authorrunning{J.Cai, T.Huang, B.Dilkina}
%
\institute{University of Southern California, Los Angeles CA, USA \\
\email{\{caijunya, taoanhua, dilkina\}@usc.edu}}
\maketitle              
\begin{abstract}
Mixed Integer Linear Programs (MILPs) are highly flexible and powerful tools for modeling and solving complex real-world combinatorial optimization problems.
Recently, machine learning (ML)-guided approaches have demonstrated significant potential in improving MILP-solving efficiency. 
However, these methods typically rely on separate offline data collection and training processes, which limits their scalability and adaptability.
This paper introduces the first multi-task learning framework for ML-guided MILP solving. The proposed framework provides MILP embeddings helpful in guiding MILP solving across solvers  (e.g., Gurobi and SCIP) and across tasks (e.g., Branching and Solver configuration).
Through extensive experiments on three widely used MILP benchmarks, we demonstrate that our multi-task learning model performs similarly to specialized models within the same distribution. Moreover, it significantly outperforms them in generalization across problem sizes and tasks.

\end{abstract}
\section{Introduction}
Many real-world \textbf{problem domains}, such as path planning~\cite{pohl1970heuristic}, scheduling~\cite{floudas2005mixed,cai2023getting}, and network design~\cite{dilkina2010solving,huang2020enhancing}, fall into the category of combinatorial optimization (CO) and are generally NP-hard to solve. Designing efficient algorithms for CO problems is both important and challenging. Mixed Integer Linear Programs (MILPs) provide a versatile framework for modeling and solving various CO problems. MILPs involve optimizing a linear objective function subject to linear constraints, with some variables restricted to integer values. Significant advancements in MILP solvers, such as Gurobi~\cite{gurobi} and SCIP~\cite{bolusani2024scip}, have been achieved by leveraging techniques like Branch-and-Bound (BnB)~\cite{land2010automatic}, complemented by a suite of heuristics to enhance performance.

Recent advancements in machine learning (ML) offer new avenues to improve MILP solvers. ML methods, by learning from complex historical distributions, can enhance both exact solvers like BnB and heuristic solvers. For exact solvers, ML techniques can predict \textbf{tasks} like which node to expand~\cite{song2018learning,labassi2022learning}, which variable to branch on~\cite{khalil2016learning,gasse2019exact,cai2024learning}, which cut to apply~\cite{tang2020reinforcement,paulus2022learning}, or how to schedule heuristics~\cite{khalil2017learning,chmiela2021learning,hendel2022adaptive}. For heuristic solvers, ML approaches can predict solutions directly~\cite{ding2020accelerating,khalil2022mip,huang2024contrastive} or integrate Large Neighborhood Search into MILP solvers~\cite{song2020general,huang2023searching}. These ML-driven enhancements hold the promise of bridging gaps in traditional solvers.

While learning-based methods for MILPs have demonstrated success in single-task settings, they face limitations in generalizing across multiple domains and tasks. Current approaches often rely on specialized models tailored to individual tasks and domains, resulting in computationally expensive training pipelines and the need for carefully curated datasets. This lack of generalization hampers real-world applicability. To address this, researchers have begun to explore single models capable of addressing multiple problem domains, drawing inspiration from foundation models. However, these efforts have largely been restricted to single-task settings~\cite{huang2024distributional,li2024towards,drakulic2024goal}. Despite these breakthroughs, an important observation remains unaddressed: MILPs from the same problem domain often share structural and characteristic similarities that can be leveraged for multi-task learning.

This paper presents a unified multi-task learning framework for MILP solving, designed to learn shared representations across multiple tasks within a problem domain. Our approach involves a two-step training process: first, we train a shared representation layer for MILPs alongside fixed, randomly initialized task-specific output layers; then, we fine-tune the task-specific layers while keeping the shared representation layer fixed. We evaluate our framework across three tasks—\textbf{Backdoors} (root-node branching)\cite{cai2024learning}, \textbf{Predict-and-Search} (PaS)\cite{huang2024contrastive}, and \textbf{Solver Configurations}—and three common problem domains: Combinatorial Auction (CA)\cite{leyton2000towards}, Maximal Independent Set (MIS)\cite{tarjan1977finding}, and Minimum Vertex Cover (MVC)~\cite{dinur2005hardness}.

Our main findings and contributions are as follows:
\begin{itemize}
    \item \textbf{Multi-task Learning Framework:} We introduce a unified multi-task learning framework for MILPs that leverages a shared representation layer and task-specific output heads, enabling efficient adaptation on unseen tasks.
    \item \textbf{Generalization on Size:} Our multi-task models achieve competitive results compared to task-specific models within the same distribution and show superior generalization on more significant problem instances.
    \item \textbf{Cross-Task Generalization:} Through cross-task evaluations, we multi-task train on two tasks and fine-tune on the third, which consistently outperforms specialized models trained on the third task.
\end{itemize}
The remainder of this paper provides background, related work, methodology, experimental results, and discussions.

\section{Background}
This section defines MILPs and provides background knowledge on our three tasks: \backdoor, \pas, and \configuration.

\subsection{Mixed Integer Linear Programming}
A Mixed Integer Linear Program (MILP) $P = (A,b,c,I)$ is defined as:
\[
\min\{c^Tx \mid Ax \leq b, \, x \in \mathbb{R}^n, \, x_j \in \{0,1\} \, \forall j \in I\}, 
\]
where $A \in \mathbb{R}^{m \times n}$, $b \in \mathbb{R}^m$, $c \in \mathbb{R}^n$, and $I \subseteq \{1, ..., n\}$ is the set of indices for binary variables. The objective is to minimize $c^Tx$ by finding a feasible assignment for $x$ that satisfies the constraints. MILP solvers rely heavily on Branch-and-Bound (BnB)~\cite{land2010automatic} that constructs a search tree to find feasible solutions with minimum costs. This process involves repeatedly solving LP relaxations of the MILP and branching on integer variables that are fractional in the LP solution, creating subproblems until all integrality constraints are met.

A key aspect of MILP solvers is their vast configuration space, with parameters spanning integer, continuous, and categorical values, influencing nearly every step of the BnB process. While the solvers' default settings aim for robust performance across heterogeneous MILP benchmarks, there is significant potential to improve configuration settings for specific distributions of instances by selecting the solver parameters effectively (\configuration).

\subsection{Backdoors for MILP}
Initially introduced for Constraint Satisfaction Problems~\cite{williams2003backdoors}, backdoors were later generalized to MILPs~\cite{dilkina2009backdoors}. In the context of MILPs, strong backdoors are defined as subsets of integer variables such that branching exclusively on these variables yields an optimal integral solution. Research~\cite{fischetti2011backdoor} has further demonstrated speedups in MILP solving times by prioritizing branching backdoor variables instead of branching exclusively on them. 

Given a MILP instance $P = (A,b,c,I)$, a pseudo-backdoor (\backdoor) of size $K \ll |I|$ is a small subset $B \subset I$ of binary variables, with $|B| = K$, whose variables are prioritized for branching to improve solver performance~\cite{ferber2022learning,cai2024learning}. We guide the solver's decision-making process by assigning higher branching priority to the variables in $B$ at the start of the tree search. This adjusted branching order influences the solver’s primal heuristics and enhances the overall pruning efficiency in the BnB procedure.

\subsection{Predict-and-Search}
Predict-and-Search (\pas)~\cite{han2023gnn} is a primal heuristic that leverages the prediction of the optimal solutions to guide the search process. Given a MILP instance, $P = (A,b,c,I)$, let \(p_\theta(x_i \mid P)\) denote the predicted probability for each binary variable \(x_i \in I\). \pas identifies near-optimal solutions by exploring a neighborhood informed by these predictions. Specifically, it selects \(k_0\) binary variables \(X_0\) with the smallest \(p_\theta(x_i \mid P)\) and \(k_1\) binary variables \(X_1\) with the largest \(p_\theta(x_i \mid P)\), ensuring \(X_0\) and \(X_1\) are disjoint (\(k_0 + k_1 \leq q\)). Variables in \(X_0\) are fixed to 0, and those in \(X_1\) are fixed to 1 in a sub-MILP. However, \pas allows up to \(\Delta \geq 0\) of these fixed variables to be flipped during solving.
Formally, let 
$
B(X_0, X_1, \Delta) = \{x : \sum_{x_i \in X_0} x_i + \sum_{x_i \in X_1} (1 - x_i) \leq \Delta \}
$
be the neighborhood defined by \(X_0\), \(X_1\), and \(\Delta\), and let \(D\) represent the feasible region of the original MILP. \pas then solves the following optimization problem:
$
\min c^T x \quad \text{s.t.} \quad x \in D \cap B(X_0, X_1, \Delta).
$
Restricting the solution space to \(D \cap B(X_0, X_1, \Delta)\) simplifies the problem, enabling the solver to find high-quality feasible solutions to the original MILP more efficiently.

\section{Related Work}
This section provides an overview of related work in learning techniques for MILP solving, focusing on learning to branch, solution prediction, and algorithm configuration. Additionally, it highlights research about generalization in the context of ML-guided solving.

\subsection{Machine Learning for MILP Solving}
There has been extensive research leveraging machine learning techniques to improve MILP solvers. Common approaches represent MILPs as bipartite graphs~\cite{gasse2019exact} and employ graph neural networks (GCN~\cite{gasse2019exact,ding2020accelerating,paulus2022learning,khalil2022mip,valentin2022instance,huang2024contrastive,hosny2024automatic} and GAT~\cite{huang2023searching,cai2024learning}) to learn various MILP decisions. Learning methods are diverse, some commonly used are imitation learning~\cite{he2014learning,song2018learning,paulus2022learning}, contrastive learning~\cite{huang2023searching,cai2024learning,huang2024contrastive,lin2024cambranch}, and reinforcement learning~\cite{tang2020reinforcement,song2020general,cai2024balans}. For a comprehensive survey on machine learning for MILP solving, we refer readers to~\cite{scavuzzo2024machine}. Here, we focus on our selected three tasks.

\textbf{Learning to branch:} Several studies~\cite{khalil2016learning,lodi2017learning,alvarez2017machine,gasse2019exact,lin2024cambranch} have explored learning to branch by imitating strong branching heuristics and predicting scores or ranking variables. Still, these approaches require solver-specific implementations and multiple test-time inferences. In contrast, backdoor approaches~\cite{ferber2022learning,cai2024learning} focus on predicting branching variables at the root node, treating the solver as a black box with a single inference. The first work to use ML-guided techniques for identifying effective backdoors is~\cite{ferber2022learning}, which employs a scorer model and a classifier model trained on data collected via biased sampling methods from~\cite{dilkina2009backdoors}. Building on this,~\cite{cai2024learning} utilizes a contrastive learning model to generate backdoors, leveraging a novel Monte Carlo tree search-based data collection approach introduced in~\cite{khalil2022finding}.

\textbf{Solution prediction:} The goal is to predict partial assignments of high-quality feasible solutions in a MILP to guide the search. \cite{ding2020accelerating,tong2024optimization} identifies variables that remain unchanged across near-optimal solutions and searches within their neighborhood. \cite{nair2020solving} and \cite{khalil2022mip} propose fixing predicted variables and letting the MILP solver optimize the rest as a warm start. However, if predictions are inaccurate, fixing variables can lead to low-quality or infeasible solutions. \cite{han2023gnn} introduces PaS, which searches for solutions within a predefined neighborhood of the prediction, improving feasibility and quality. \cite{huang2024contrastive} extends PaS using contrastive learning and novel optimization-based methods for handling low-quality or infeasible solutions.

\textbf{Instance-Specific Configuration:} Introduced by~\cite{kadioglu2010isac}, this approach extracts features from problem instances and uses G-means clustering to group similar instances for configuration selection. Hydra-MIP~\cite{xu2011hydra} improved this by incorporating features from short solver runs before selecting configurations for full runs. In the NeurIPS 2021 ML4CO competition~\cite{gasse2022machine}, participants successfully applied ML-guided regression methods to choose the best configurations~\cite{valentin2022instance}. \cite{hosny2024automatic} learns MILP similarities related to solution costs and uses K-nearest neighbors to select configurations at inference time. Different from choosing the best from a set of configurations, we propose a contrastive learning approach that learns to generate a new configuration by learning to discriminate between good and bad ones.

\subsection{ML-guided solving Generalization}
Despite advancements in ML-guided solving, generalizing learned models across tasks and problem domains remains a key challenge. Most ML methods are tailored to specific problem classes, limiting their applicability in real-world scenarios. Recently, some works have addressed this issue. \cite{huang2024distributional} introduced Distributional MIPLIB, the first multi-domain library for advancing ML-guided MILP methods and exploring cross-domain generalization. \cite{li2024towards} proposed MILP-Evolve, leveraging large language models to generate diverse MILP classes, showing strong generalization when trained on a large dataset. \cite{drakulic2024goal} introduced GOAL, a generalist model for various combinatorial optimization problems, but it has not been applied to MILP domains. Additionally, multi-task learning has been applied in other CS problems like computer vision~\cite{liu2019end} and natural language processing~\cite{chen2023minigpt}, but no one has applied it to the CO domain.

\begin{figure*} [ht]
    \centering
    \includegraphics[width=\linewidth]{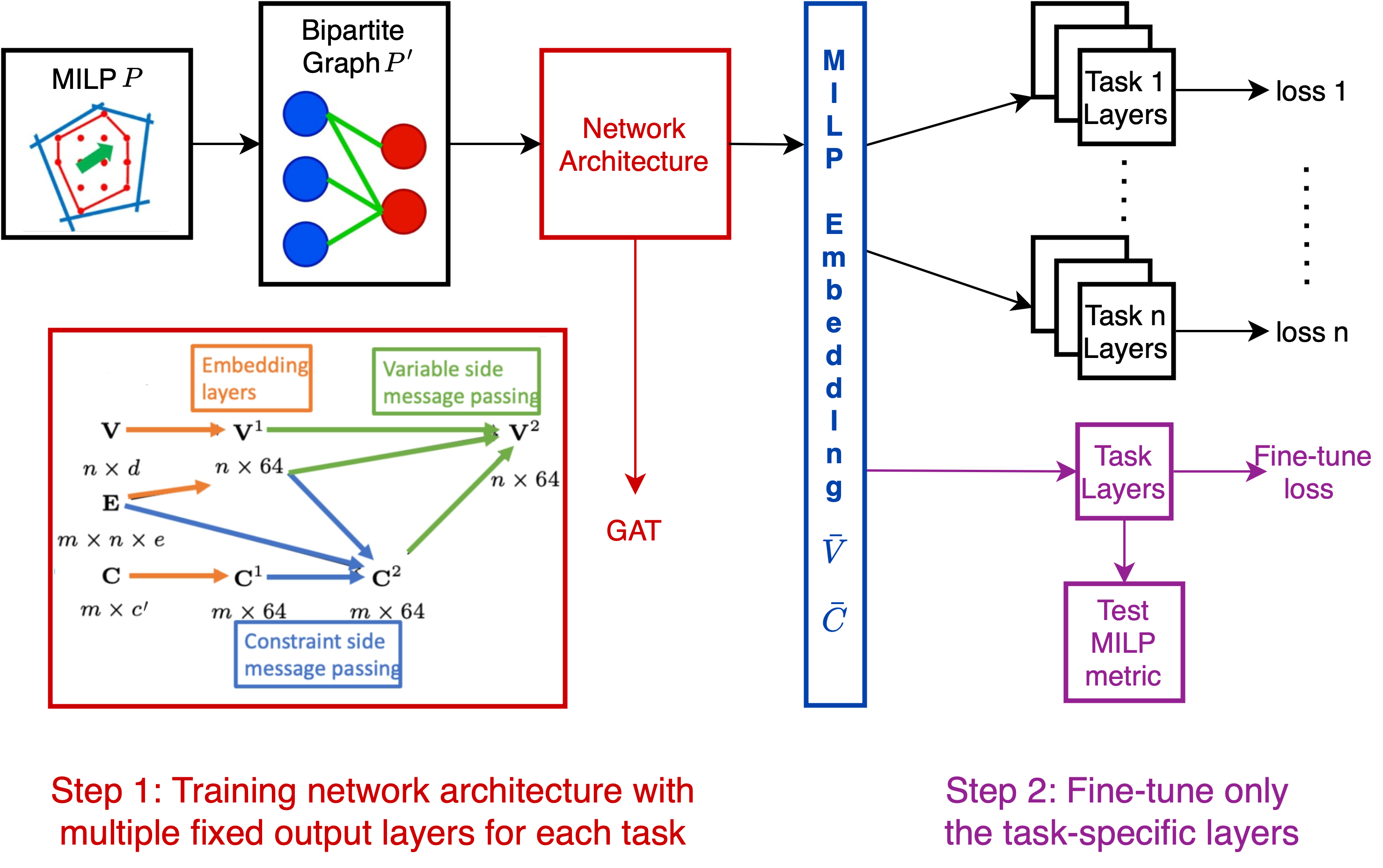}
    \caption{This figure illustrates the multi-task learning framework for Mixed Integer Linear Programs (MILPs). Step 1 (red) involves training a network architecture with a graph representation of MILPs to produce general MILP embeddings for multiple random task-specific layers. In Step 2 (purple), task-specific layers are fine-tuned for different downstream tasks, optimizing task-specific losses while preserving a shared representation and evaluating using task-specific metrics. In our settings,  MILP embedding $(\bar{V}, \bar{C}) = (V^2, C^2)$ and number of tasks $n=2$.}
    \label{fig: pipeline}
\end{figure*}

\section{Multi-task Representation Learning}
We aim to learn MILP embeddings that are effective across different learning tasks and instances within the same problem domain, and that can be easily fine-tuned for new tasks. We share the same network architecture for different tasks to enable multi-task training with task-specific layers attached. Training data from other tasks is processed alternately by batch in every epoch.

However, this alternating training strategy often leads to competition between different tasks, causing significant oscillations in the loss curves and a strong bias in testing performance towards one of the tasks. To address this issue, we introduced a two-step training strategy. First, we train the shared network architecture while keeping the task-specific layers fixed at randomly initialized weights. We use three randomly initialized task-specific layers to enhance the robustness of each task. This ensures the shared network architecture learns a general MILP embedding that is not overly specialized to a single random initialization. Second, we fine-tune the task-specific layers for each task while keeping the shared network architecture fixed. This step ensures that the standard MILP embedding benefits each specific task without disrupting the shared representation.

The pipeline is shown in Figure \ref{fig: pipeline}. With this multi-task training pipeline, we achieve a standard representation embedding of MILPs that can be generalized across different tasks within the same problem domain. It is worth noting that this pipeline is adaptable to \textbf{any graph representation of MILPs}, \textbf{any size-invariant network architecture}, and \textbf{any loss function}. This paper focuses on the bipartite graph representation, Graph Attention Networks, and contrastive loss. In the following subsections, we explain the data collection process, bipartite graph representation, network architecture, loss function, and applying learned networks for the tasks: \backdoor, \pas, and \configuration. We choose these three MILP tasks because there are existing works~\cite{cai2024learning,huang2024contrastive} that have utilized the same architecture and loss function, which allows us to effectively explore a multitask model without starting entirely from scratch. Additionally, these three tasks require only one-time inference with static features.

\subsection{Data collection and Data representation}
Since we use contrastive loss, carefully collecting positive and negative samples for practical training is crucial. For \backdoor, we follow~\cite{cai2024learning} and employ the Monte Carlo Tree Search (MCTS) algorithm proposed in~\cite{khalil2022finding} to generate candidate backdoors. Positive samples are selected as the backdoors with the shortest runtimes, while negative samples are chosen as the backdoors with the longest runtimes. For \configuration, we use SMAC3~\cite{lindauer2022smac3} to collect candidate configurations. Positive samples are the configurations with the best feasible solution found, while negative samples are the configurations with the worst feasible solution found. For \pas, we follow~\cite{huang2024contrastive} and collect a set of optimal or near-optimal solutions as positive samples. We identify low-quality solutions for negative samples by finding the worst feasible solutions that differ from the positive samples in, at most, 10\% of the binary variables.

We represent the MILP $P = (A, b, c, I)$ as a bipartite graph, following the approach in~\cite{gasse2019exact}. The resulting bipartite graph, denoted as $P' = (G, V, C, E)$, consists of a graph $G$ with two types of nodes: variable nodes and constraint nodes. An edge $(i, j)$ exists between a variable node $i$ and a constraint node $j$ if the variable $i$ appears in the constraint $j$ with a nonzero coefficient, i.e., $A_{ji} \neq 0$. Constraint, variable, and edge features are represented as matrices $V \in \mathbb{R}^{n \times d}$, $C \in \mathbb{R}^{m \times c'}$, and $E \in \mathbb{R}^{m \times n \times e}$, respectively. This bipartite graph representation ensures that the MILP encoding is invariant to permutations of variables and constraints. Additionally, it enables the use of predictive models designed for graphs of varying sizes, allowing deployment on problems with differing numbers of variables and constraints. The features utilized include 15 variable features (e.g., variable types, coefficients, upper and lower bounds, and root LP-related features), four constraint features (e.g., constant terms and senses), and one edge feature (the coefficients).

\subsection{Network architecture and Contrastive loss}
Our network architecture is a Graph Attention Network (GAT)~\cite{brody2021attentive}, which takes the bipartite graph $P'$ as input and outputs the MILP embeddings $(V^2, C^2)$. To enhance the modeling capacity and manage interactions between nodes, embedding layers are employed to adjust the sizes of the feature embeddings to $V^1 \in \mathbb{R}^{n \times L}$ and $C^1 \in \mathbb{R}^{m \times L}$. Subsequently, the GAT performs two rounds of message passing. In the first round, each constraint node in $C^1$ attends to its neighboring variable nodes via an attention mechanism with $H$ attention heads, producing updated constraint embeddings $C^2$. Similarly, each variable node in $V^1$ attends to its neighboring constraint nodes in the second round, yielding updated variable embeddings $V^2$ using a separate set of attention weights. The GAT is designed to learn a shared MILP embedding representation $(V^2, C^2)$ that is generalizable across various tasks and instances. In the experiments, embedding vector size $L$ and number of attention heads $H$ are set to 64 and 8. 

The task-specific layers vary depending on the task. For \backdoor and \pas tasks, only variable features are required during prediction. In these cases, the task-specific layers consist of a multi-layer perceptron with a sigmoid activation function, which outputs a score between 0 and 1 for each variable. For \configuration, we have additional layers combining variable and constraint features, mapping the embedding size to the number of configuration parameters, and finally, sigmoid and softmax functions are used to generate scores between 0 and 1 for numerical and categorical parameters.

The model is trained or fine-tuned using a contrastive loss function that scores parameters by learning to emulate superior samples while avoiding inferior ones. Given a set of MILP instances $\mathcal{P}$ for training, let $\mathcal{D} = \{(S^P_+, S^P_-) : P \in \mathcal{P}\}$ represent the set of positive and negative samples for all training instances. Let $p_\theta(P)$ denote the model’s prediction for an instance $P$ with network parameters $\theta$. Using dot products for similarity, the InfoNCE~\cite{oord2018representation} contrastive loss is defined as:
\[
\mathcal{L}(\theta) = \sum_{(S^P_+, S^P_-) \in \mathcal{D}} \frac{-1}{|S^P_+|} \sum_{a \in S^P_+} \log \frac{\exp(a^\top p_\theta(P) / \tau)}{\sum\limits_{a' \in S^P_- \cup \{a\}} \exp(a'^\top p_\theta(P) / \tau)},
\]
where $\tau$ is a temperature hyperparameter set to $0.07$ in the experiments, following~\cite{huang2023searching}. 

\subsection{Applying learned network}
During testing, given a MILP instance, we convert it to a bipartite graph and inference one-time with the network to output a score vector with one score for each variable/parameter. 
\begin{itemize}
    \item For \backdoor, the binary variables with the highest scores are greedily selected as the predicted backdoors based on a user-defined backdoor size $K$.
    \item For \pas, \(X_0\) and \(X_1\) are selected greedily based on the predictions, and a constrained optimization problem is solved using the hyperparameters $k_0$, $k_1$, and $\Delta$.
    \item For \configuration, categorical parameters are set to the option with the highest score, while numerical parameters use the output score.
\end{itemize}
Full details of bipartite graph features, GAT network architecture, and hyperparameter settings are provided in the Appendix. 

\section{Experiments}
This section introduces the setup for empirical evaluation and presents the results. The code and Appendix are available at \url{https://github.com/caidog1129/MILP_multitask}.

\begin{table}[t]
\centering
\caption{Instance sizes for the benchmarks: Combinatorial Auctions (CA), Maximum Independent Set (MIS), and Minimum Vertex Cover (MVC) across the tasks \backdoor, \pas, and \configuration. The table distinguishes between small (S) and large (L) instances. For CA, the instance sizes are characterized by the number of items and bids. For MIS and MVC, the instance sizes are described by the average degree and the number of nodes.}
\begin{tabular}{c|c|c|c|c}
\toprule
Benchmarks & Description                           & \backdoor  & \pas        & \configuration \\ \midrule
CA-S       & \multirow{2}{*}{(\# items, \# bids)}    & (175, 850)  & (2000, 4000) & (2000, 4000)    \\
CA-L       &                                       & (200, 1000) & (3000, 6000) & (3000, 6000)    \\ \midrule
MIS-S      & \multirow{2}{*}{(avg degree, \# nodes)} & (4, 1250)   & (5, 6000)    & (4, 3000)       \\
MIS-L      &                                       & (4, 1500)   & (5, 9000)    & (5, 6000)       \\ \midrule
MVC-S      & \multirow{2}{*}{(avg degree, \# nodes)} & (5, 1500)   & (5, 6000)    & (4, 3000)       \\
MVC-L      &                                       & (5, 2000)   & (5, 9000)    & (5, 6000)       \\
\bottomrule
\end{tabular}
\label{table: instances}
\end{table}

\subsection{Experiment Setup}
\textbf{Benchmark Problems and Instance Generation:} \\
We evaluate our approach on three NP-hard benchmark problems widely used in existing studies~\cite{gasse2019exact,han2023gnn}: combinatorial auction (CA), minimum vertex cover (MVC), and maximum independent set (MIS). Previous works~\cite{cai2024learning,huang2024contrastive} have demonstrated promising results in predicting backdoor variables and PaS assignments on these problem domain benchmarks. MVC and MIS instances are generated using the Barabási–Albert random graph model~\cite{albert2002statistical}, while CA instances are generated based on arbitrary relations described in~\cite{leyton2000towards}. This paper focuses on the formulation that includes only binary variables as it aligns with the common problem domains in our three tasks. Still, our multi-task learning approach works on the general MILP problem domain.

We utilize Distributional-MIPLIB~\cite{huang2024distributional} to generate 200 training small (S) instances and 100 test S instances for each task. We also generated another 100 test large (L) instances for each task to test the model's generalizability. We train and fine-tune the model on S instances and test on both S and L instances. Each task involves different problem sizes: \backdoor focuses on improving optimal solving time for smaller instances (with optimal solutions found in approximately hundreds of seconds), while \pas and \configuration address more challenging instances (solving optimally takes over an hour), aiming to achieve better primal solutions within a runtime cutoff. Table~\ref{table: instances} provides detailed information on the generated instances for each problem domain and task. \\

\noindent \textbf{Baselines and Approaches:} \\
We compare our approach with baseline solvers: \gurobi~\cite{gurobi} (default settings) for \backdoor and \pas, and \scip~\cite{bolusani2024scip} (default configuration) for \configuration. \gurobi, the state-of-the-art commercial solver, and \scip, the best open-source solver, provide a diverse evaluation framework, demonstrating that our multi-task learning framework is not tied to a single solver.

For \configuration, we include an additional baseline, \smac, which uses SMAC3~\cite{lindauer2022smac3} to perform configuration space search in 5 rounds per test instance. We do not choose \gurobi\ for \configuration because we cannot find a set of working configurations from previous literature or our experiments. We do not compare with other learning-based methods~\cite{hosny2024automatic}, as their approach requires an extensive database of pre-explored configurations to select similar instances, whereas our task generates new configurations specific to each instance. 

The model trained specifically on a single task is denoted as \singletask. We have three single-task models, each focusing on a specific task. For simplicity, we use consistent notation to indicate which task each model is associated with. On the other hand, the model trained using our multi-task learning framework is referred to as \textit{Multi-task} (\multitaskbapas, \multitaskbaconfig, and \multitaskpasconfig), where the suffix indicates the two tasks used in multi-task training). For \singletask, we train on 200 instances for each task. For \textit{Multi-task}, we first train on 200 instances from each task with multiple random task-specific layers fixed; then, we fine-tune the task-specific layer from scratch on the same 200 instances used in the first step. For new tasks, we use the same 200 instances for both \singletask training and fine-tuning new task-specific layers. \\

\noindent \textbf{Evaluation Metrics:} \\
We evaluate performance on 100 test instances using the following metrics:
\begin{enumerate}
    \item \textbf{Solve Time:} The time in seconds required for the solver to find the optimal solution to the MILP. This metric evaluates the efficiency of solving optimization problems.
    \item \textbf{Primal Gap (PG):}~\cite{berthold2006primal} Defined as the normalized difference between the primal bound $v$ and the best-known objective value $v^*$:
    \[
    \text{PG} = \frac{|v - v^*|}{\max(|v^*|, \epsilon)},
    \]
    where $\epsilon = 10^{-8}$ avoids division by zero. This metric applies when $v$ exists and $v v^* \geq 0$.
    \item \textbf{Primal Integral (PI):}~\cite{achterberg2012rounding} The integral of the primal gap over runtime $[0, t]$. PI reflects both the quality and speed of finding solutions.
\end{enumerate}

We evaluate solve time for the \backdoor, while for \pas and \configuration, we measure the primal gap and primal integral within a runtime cutoff. We do not collect primal gap or primal integral for the \backdoor, as doing so would interfere with the solving process, particularly for instances that can be solved in hundreds of seconds. \\

\begin{table}[t]
\centering
\caption{\textbf{Same-Task Performance:} Solve time (seconds) for \backdoor and Primal Integral for \pas averaged over 100 test instances for each benchmark. We compare the performance of \gurobi, \singletask (trained on \backdoor or \pas), and \multitaskbapas. Results include the mean, standard deviation, and the number of instances each approach wins. The best-performing entries are highlighted in bold for clarity.\\}
\begin{tabular}{c|c|ccc|ccc}
\toprule
                       &                  & \multicolumn{3}{c|}{\backdoor Solve Time}                   & \multicolumn{3}{c}{\pas Primal Integral}                      \\ \midrule
Benchmarks             & Approaches       & Mean            & Std Dev         & Wins        & Mean          & Std Dev       & Wins         \\ \midrule
\multirow{3}{*}{CA-S}  & \gurobi           & 252.66          & 122.06          & 5           & 14.27         & 5.28          & 8            \\
                       & \singletask      & 228.46          & \textbf{114.83} & 44          & 9.43          & 5.11          & 45           \\
                       & \multitaskbapas & \textbf{219.23} & 115.13          & \textbf{51} & \textbf{9.36} & \textbf{4.99} & \textbf{47}  \\ \midrule
\multirow{3}{*}{MIS-S} & \gurobi            & 150.98          & 204.42          & 29          & 16.33         & 4.01          & 0            \\
                       & \singletask      & 141.76          & 204.02          & \textbf{39} & \textbf{3.00} & 1.18          & \textbf{59}  \\
                       & \multitaskbapas & \textbf{133.18} & \textbf{190.68} & 32          & 3.35          & \textbf{1.16} & 41           \\ \midrule
\multirow{3}{*}{MVC-S} & \gurobi            & 81.06           & \textbf{113.24} & 7           & 1.27          & 0.60          & 0            \\
                       & \singletask      & 75.83           & 122.63          & \textbf{53} & \textbf{0.52} & \textbf{0.09} & 43           \\
                       & \multitaskbapas & \textbf{74.65}  & 113.81          & 40          & \textbf{0.52} & 0.12          & \textbf{57}  \\ \midrule
\multirow{3}{*}{CA-L}  & \gurobi            & 831.43          & 464.74          & 20          & 19.53         & 6.70          & 7            \\
                       & \singletask      & 784.77          & 511.40          & 33          & 14.14         & 5.86          & 24           \\
                       & \multitaskbapas & \textbf{740.08} & \textbf{409.45} & \textbf{47} & \textbf{9.88} & \textbf{4.88} & \textbf{68}  \\ \midrule 
\multirow{3}{*}{MIS-L} & \gurobi            & 489.25          & 840.36          & 22          & 30.73         & 8.34          & 0            \\
                       & \singletask      & 433.18          & 734.15          & 32          & 50.83         & 8.27          & 0            \\
                       & \multitaskbapas & \textbf{401.23} & \textbf{666.22} & \textbf{46} & \textbf{2.23} & \textbf{0.74} & \textbf{100} \\ \midrule                       
\multirow{3}{*}{MVC-L} & \gurobi            & 461.59          & 1062.18         & 30          & 3.44          & 1.88          & 0            \\
                       & \singletask      & 413.63          & 735.18          & 29          & 1.19          & 0.52          & 30           \\
                       & \multitaskbapas & \textbf{393.69} & \textbf{711.34} & \textbf{41} & \textbf{0.98} & \textbf{0.40} & \textbf{70}  \\ \bottomrule
\end{tabular}
\label{table: backdoorpas}
\end{table}

\noindent \textbf{Hyperparameters:} \\
Experiments are conducted on 2.4 GHz Xeon-2640v3 CPUs with 64 GB memory. Training is performed on an NVIDIA V100 GPU with 112 GB memory. We use Gurobi 10.0.2 and SCIP 9.0.0 as solvers. For training and fine-tuning, the Adam optimizer~\cite{kingma2014adam} is used with a learning rate of $1 \times 10^{-4}$. The batch size is set to 32, and training and fine-tuning are run for 1000 epochs (training converges in less than 12 hours, and fine-tuning converges in less than 1 hour). The number of parameters in the shared layers and task-specific layers is roughly a ratio of 10:1. During testing, the model with the best validation loss is selected. Runtime cutoffs are set to 1000 seconds for \pas and 900 seconds for \configuration instances. For the backdoor task, we follow~\cite{cai2024learning} to select backdoor size $K$. For the \pas task, we use $k_0$, $k_1$, and $\Delta$ as described in~\cite{huang2024contrastive}. For the \configuration task, we select 15 parameters related to the solving process (e.g., branching, cut selection, LP relaxation) based on SCIP 9.0.0, following~\cite{hosny2024automatic}.

Additional details on instance generation, data collection, configuration parameters, and hyperparameter settings are provided in the Appendix. \\

\begin{figure}[t]
\includegraphics[width=\textwidth]{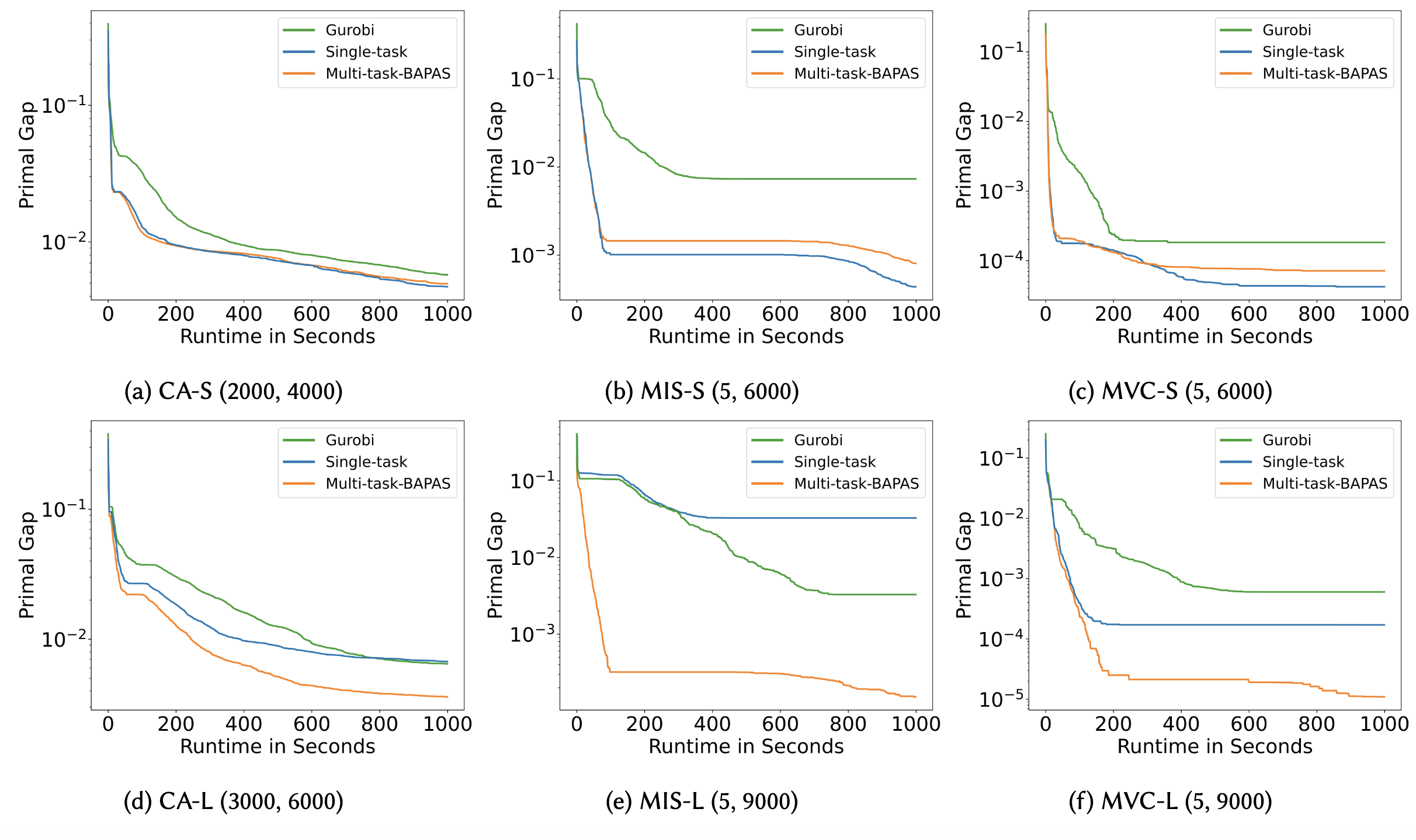}
 \caption{\textbf{Same-Task Performance:} The Primal Gap (the lower, the better) as a runtime function averaged over 100 test instances on \pas and each benchmark. We compare the performance of \gurobi\ (green line), \singletask on \pas (blue line), and \multitaskbapas (orange line).}
 \label{fig: pas}
\end{figure}

\noindent \textbf{Our experiments aim to answer the following research questions:}
\begin{itemize}
\item Does our newly proposed approach for learning to generate configurations for MILP solvers outperform default solvers and other baseline methods?
\item How does \textit{Multi-task} compare to \singletask and default solvers in performance on the same task? Additionally, how well does it generalize to larger instance sizes?
\item Can the model, with its shared MILP embeddings, effectively fine-tune task-specific layers to handle new tasks, and how does its performance compare to both \singletask and default solvers?
\item Does \textit{Multi-task} consistently outperform \singletask in terms of generalization, as demonstrated through cross-task evaluations?
\end{itemize}

\begin{figure}[t]
\includegraphics[width=\textwidth]{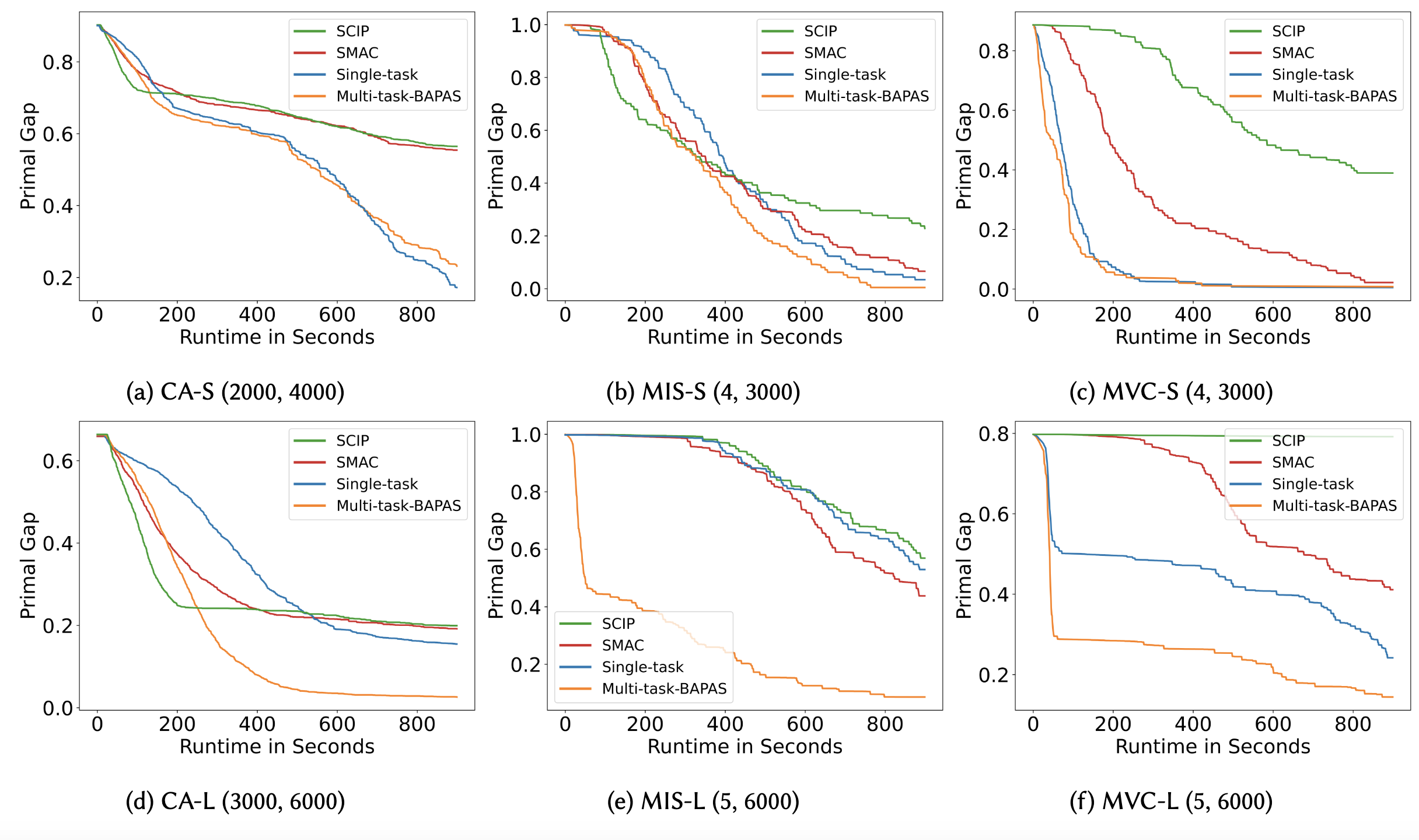}
 \caption{\textbf{New-task Performance:} The Primal Gap (the lower, the better) as a runtime function averaged over 100 test instances on \configuration and each benchmark. We compare the performance of \scip\ (green line), \smac\ (red line), \singletask on \configuration (blue line), and \multitaskbapas (orange line).}
 \label{fig: config}
\end{figure}

\subsection{Results}
\noindent \textbf{Same-Task Performance:} \\
To evaluate the same-task performance of a multi-task model, we test it on the tasks on which it is trained. We evaluate \gurobi, \singletask, and \multitaskbapas on \backdoor and \pas using both S and L instances from CA, MIS and MVC. The detailed results are shown in Table \ref{table: backdoorpas} and Figure \ref{fig: pas}. For 100 S instances, both \singletask and \multitaskbapas outperform \gurobi\ consistently, demonstrating competitive performance. For \backdoor, \multitaskbapas achieves a slightly better average runtime than \singletask, with improvements of 4.04\%, 6.05\%, and 1.56\% in CA-S, MIS-S, and MVC-S, respectively. However, \singletask outperforms \multitaskbapas in the number of instances, with 39 vs. 32 wins in MIS-S and 53 vs. 40 in MVC-S. For \pas, while \singletask consistently achieves a better primal gap at the end of the time cutoff, \multitaskbapas demonstrates slightly better performance in terms of the primal integral, securing 2 and 14 more wins in CA-S and MVC-S, respectively.

The results differ significantly when evaluating generalization performance on 100 L instances (directly tested without training). The multi-task model substantially outperforms both \singletask and \gurobi. For \backdoor, the multi-task model achieves the best average runtime across all benchmarks, showing an approximately 15\% improvement over \gurobi, compared to a 9\% improvement by \singletask. The multi-task model also secures more than 40 wins per benchmark. Additionally, a shift in the distribution of wins is observed, with \gurobi\ achieving a more significant number of wins and \singletask showing a reduced number of wins compared to the multi-task model. For \pas, \multitaskbapas demonstrates a significantly better primal gap across all benchmarks at the end of the runtime cutoff. While the \singletask initially performs better than \gurobi\ on CA-L, it is eventually surpassed by \gurobi\ at the end of the runtime. Additionally,  \singletask performs poorly on MIS-L, even underperforming compared to \gurobi. In contrast, \multitaskbapas, trained on the same data, avoids this issue entirely. Regarding the primal integral, \multitaskbapas achieves improvements of 49.42\%, 92.75\%, and 71.51\% over \gurobi\ and secures wins in most instances, further highlighting its superior performance.

\begin{table}[t]
\centering
\caption{\textbf{Cross-task Evaluations:} Solve time (seconds) for \backdoor and Primal Integral for \pas and \configuration averaged over 100 test instances for each benchmark. We compare the performance of \gurobi, \singletask, and correspond \textit{Multi-task}. Results include the mean and the number of instances each approach wins. The number of Wins does not add up to 100 for \configuration because \smac\ wins in some instances. The reforming entries are highlighted in bold for clarity. \\}
\begin{tabular}{c|c|cc|cc|cc}
\toprule
                                                                                           &                     & \multicolumn{2}{c|}{CA-L}     & \multicolumn{2}{c|}{MIS-L}     & \multicolumn{2}{c}{MVC-L}     \\ \midrule
Tasks                                                                                      & Approaches          & Mean            & Wins        & Mean            & Wins         & Mean            & Wins        \\ \midrule
\multirow{3}{*}{\begin{tabular}[c]{@{}c@{}}\backdoor \\ Solve Time\end{tabular}}          & \gurobi             & 793.10          & 17          & 530.41          & 25           & 427.86          & 20          \\
                                                                                           & \singletask         & 749.08          & 28          & 470.73          & 31           & 390.04          & 24          \\
                                                                                           & \multitaskpasconfig & \textbf{696.39} & \textbf{55} & \textbf{446.61} & \textbf{44}  & \textbf{354.35} & \textbf{56} \\ \midrule
\multirow{3}{*}{\begin{tabular}[c]{@{}c@{}}\pas \\ Primal Integral\end{tabular}}          & \gurobi              & 19.97           & 8           & 37.95           & 0            & 3.84            & 0           \\
                                                                                           & \singletask         & 14.11           & 32          & 56.11           & 0            & 1.49            & 10          \\
                                                                                           & \multitaskbaconfig  & \textbf{12.12}  & \textbf{60} & \textbf{3.13}   & \textbf{100} & \textbf{1.04}   & \textbf{90} \\ \midrule
\multirow{3}{*}{\begin{tabular}[c]{@{}c@{}}\configuration\\ Primal Integral\end{tabular}} & \scip               & 247.97          & 6           & 782.65          & 1            & 714.79          & 0           \\
                                                                                           & \singletask         & 306.77          & 4           & 768.61          & 4            & 398.38          & 21          \\
                                                                                           & \multitaskbapas     & \textbf{162.21} & \textbf{77} & \textbf{234.45} & \textbf{90}  & \textbf{231.92} & \textbf{71} \\ \bottomrule
\end{tabular}
\label{table: cross}
\end{table}

This result aligns with the understanding that \singletask are highly specialized for the specific distribution of the instances they are trained on, leading to optimal performance within the same data distribution. However, our \multitaskbapas demonstrate competitive performance across most benchmarks, with only slight underperformance in some instances. When it comes to generalization, the benefits of multi-task training become apparent. By learning a generalized embedding that avoids overfitting to a single distribution, \multitaskbapas achieves significantly better performance, leading to superior performance on unseen instances where \singletask struggles.

\noindent \textbf{New-Task Performance:} \\We compare \scip, \smac, \singletask, and \multitaskbapas on \configuration using S and L instances from CA, MIS and MVC. Results are presented in Figure \ref{fig: config} and the table with the primal integral is available in the Appendix. First, in our new task of learning to generate configurations, \singletask outperforms in the primal gap over baselines on CA-S and MVC-S, and also achieving 18.27\% and 66.97\% improvements in the primal integral over \smac. While it initially generates worse solutions than the baselines on MIS-S, it surpasses them by the end of the runtime cutoff. This demonstrates that our new task can effectively learn to generate configurations that enhance solving performance. Now focusing on \multitaskbapas, we fine-tune only the task-specific layer using the same 200 configuration instances as the \singletask, while keeping the rest of the network architecture trained on \backdoor and \pas. Surprisingly, the multi-task model performs similarly to, or even better than, the \singletask on S instances. \multitaskbapas achieves better solution progress within the first 600 seconds on CA-S and MVC-S and consistently outperforms the \singletask throughout the runtime on MIS-S. For primal integral, \multitaskbapas reaches an average of 11.3\% over \singletask across three S benchmarks.

Considering generalization performance on L instances, \singletask struggles, performing consistently worse than \smac\ on CA-L and MIS-L, possibly indicating that this task is harder to generalize than others. In contrast, \multitaskbapas performs much better than other approaches, achieving the best primal gap and primal integral, particularly excelling on CA-L and MIS-L, where the \singletask model underperforms in the primal gap. For the primal integral, \multitaskbapas shows substantial improvements over \singletask with gains of 47.13\%, 69.49\%, and 41.78\% on CA-L, MIS-L, and MVC-L, respectively. Additionally, the win rate of \multitaskbapas increases significantly when transitioning from S to L instances, rising from 44 to 77 on CA, 23 to 90 on MIS, and 59 to 71 on MVC. These experiments highlight the effectiveness of our multi-task learning framework. By fine-tuning only the task-specific layer while leveraging embeddings from two prior tasks, \multitaskbapas matches or outperforms \singletask, especially excelling in generalization to larger instances.\\

\noindent \textbf{Cross-task Evaluations:} \\
To further validate the robustness of our approach, we performed cross-task evaluations on the learned MILP embeddings against \singletask. Specifically, we evaluated \multitaskpasconfig on \backdoor, \multitaskbaconfig on \pas, and \multitaskbapas on \configuration, assessing their generalization performance on 100 large instances from each benchmark. Table \ref{table: cross} highlights consistent improvements of the multi-task models over the solver and \singletask across all benchmarks. These experiments demonstrate that the success of our multi-task training framework is not limited to a single pair of tasks but is broadly effective. Additional tables and figures for S and L instances and comparisons among different multi-task models on the same task are included in the Appendix.

\section{Conclusion}
This paper introduced a multi-task learning framework for MILP, unifying diverse tasks through shared representations and task-specific fine-tuning. Our approach demonstrated competitive performance with specialized models and significantly improved generalization across problem sizes and functions. Future work includes integrating tasks with dynamic features and extending the framework to different problem domains, advancing toward general foundation models for MILP optimization.

\paragraph{Acknowledgments}
We would like to thank the anonymous reviewers for their constructive feedback to improve this paper. The National Science Foundation (NSF) supported the research under grant number 2112533: "NSF Artificial Intelligence Research Institute for Advances in Optimization (AI4OPT)".

%
%
%
\newpage
\bibliographystyle{splncs04}
\bibliography{ref}

\begin{thebibliography}{10}
\providecommand{\url}[1]{\texttt{#1}}
\providecommand{\urlprefix}{URL }
\providecommand{\doi}[1]{https://doi.org/#1}

\bibitem{achterberg2012rounding}
Achterberg, T., Berthold, T., Hendel, G.: Rounding and propagation heuristics for mixed integer programming. In: Operations Research Proceedings 2011: Selected Papers of the International Conference on Operations Research (OR 2011), August 30-September 2, 2011, Zurich, Switzerland. pp. 71--76. Springer (2012)

\bibitem{albert2002statistical}
Albert, R., Barab{\'a}si, A.L.: Statistical mechanics of complex networks. Reviews of modern physics  \textbf{74}(1), ~47 (2002)

\bibitem{alvarez2017machine}
Alvarez, A.M., Louveaux, Q., Wehenkel, L.: A machine learning-based approximation of strong branching. INFORMS Journal on Computing  \textbf{29}(1),  185--195 (2017)

\bibitem{berthold2006primal}
Berthold, T.: Primal heuristics for mixed integer programs. Ph.D. thesis, Zuse Institute Berlin (ZIB) (2006)

\bibitem{bolusani2024scip}
Bolusani, S., Besan{\c{c}}on, M., Bestuzheva, K., Chmiela, A., Dion{\'\i}sio, J., Donkiewicz, T., van Doornmalen, J., Eifler, L., Ghannam, M., Gleixner, A., et~al.: The scip optimization suite 9.0. arXiv preprint arXiv:2402.17702  (2024)

\bibitem{brody2021attentive}
Brody, S., Alon, U., Yahav, E.: How attentive are graph attention networks? arXiv preprint arXiv:2105.14491  (2021)

\bibitem{cai2024learning}
Cai, J., Huang, T., Dilkina, B.: Learning backdoors for mixed integer programs with contrastive learning. arXiv preprint arXiv:2401.10467  (2024)

\bibitem{cai2024balans}
Cai, J., Kadioglu, S., Dilkina, B.: Balans: Multi-armed bandits-based adaptive large neighborhood search for mixed-integer programming problem. arXiv preprint arXiv:2412.14382  (2024)

\bibitem{cai2023getting}
Cai, J., Nguyen, K.N., Shrestha, N., Good, A., Tu, R., Yu, X., Zhe, S., Serra, T.: Getting away with more network pruning: From sparsity to geometry and linear regions. In: International Conference on Integration of Constraint Programming, Artificial Intelligence, and Operations Research. pp. 200--218. Springer (2023)

\bibitem{chen2023minigpt}
Chen, J., Zhu, D., Shen, X., Li, X., Liu, Z., Zhang, P., Krishnamoorthi, R., Chandra, V., Xiong, Y., Elhoseiny, M.: Minigpt-v2: large language model as a unified interface for vision-language multi-task learning. arXiv preprint arXiv:2310.09478  (2023)

\bibitem{chmiela2021learning}
Chmiela, A., Khalil, E., Gleixner, A., Lodi, A., Pokutta, S.: Learning to schedule heuristics in branch and bound. Advances in Neural Information Processing Systems  \textbf{34},  24235--24246 (2021)

\bibitem{dilkina2010solving}
Dilkina, B., Gomes, C.P.: Solving connected subgraph problems in wildlife conservation. In: CPAIOR. vol.~6140, pp. 102--116. Springer (2010)

\bibitem{dilkina2009backdoors}
Dilkina, B., Gomes, C.P., Malitsky, Y., Sabharwal, A., Sellmann, M.: Backdoors to combinatorial optimization: Feasibility and optimality. In: Integration of AI and OR Techniques in Constraint Programming for Combinatorial Optimization Problems: 6th International Conference, CPAIOR 2009 Pittsburgh, PA, USA, May 27-31, 2009 Proceedings 6. pp. 56--70. Springer (2009)

\bibitem{ding2020accelerating}
Ding, J.Y., Zhang, C., Shen, L., Li, S., Wang, B., Xu, Y., Song, L.: Accelerating primal solution findings for mixed integer programs based on solution prediction. In: Proceedings of the aaai conference on artificial intelligence. vol.~34, pp. 1452--1459 (2020)

\bibitem{dinur2005hardness}
Dinur, I., Safra, S.: On the hardness of approximating minimum vertex cover. Annals of mathematics pp. 439--485 (2005)

\bibitem{drakulic2024goal}
Drakulic, D., Michel, S., Andreoli, J.M.: Goal: A generalist combinatorial optimization agent learner. arXiv e-prints pp. arXiv--2406 (2024)

\bibitem{ferber2022learning}
Ferber, A., Song, J., Dilkina, B., Yue, Y.: Learning pseudo-backdoors for mixed integer programs. In: International Conference on Integration of Constraint Programming, Artificial Intelligence, and Operations Research. pp. 91--102. Springer (2022)

\bibitem{fischetti2011backdoor}
Fischetti, M., Monaci, M.: Backdoor branching. In: International Conference on Integer Programming and Combinatorial Optimization. pp. 183--191. Springer (2011)

\bibitem{floudas2005mixed}
Floudas, C.A., Lin, X.: Mixed integer linear programming in process scheduling: Modeling, algorithms, and applications. Annals of Operations Research  \textbf{139},  131--162 (2005)

\bibitem{gasse2022machine}
Gasse, M., Bowly, S., Cappart, Q., Charfreitag, J., Charlin, L., Ch{\'e}telat, D., Chmiela, A., Dumouchelle, J., Gleixner, A., Kazachkov, A.M., et~al.: The machine learning for combinatorial optimization competition (ml4co): Results and insights. In: NeurIPS 2021 competitions and demonstrations track. pp. 220--231. PMLR (2022)

\bibitem{gasse2019exact}
Gasse, M., Ch{\'e}telat, D., Ferroni, N., Charlin, L., Lodi, A.: Exact combinatorial optimization with graph convolutional neural networks. Advances in neural information processing systems  \textbf{32} (2019)

\bibitem{gurobi}
{Gurobi Optimization, LLC}: {Gurobi Optimizer Reference Manual} (2024), \url{https://www.gurobi.com}

\bibitem{han2023gnn}
Han, Q., Yang, L., Chen, Q., Zhou, X., Zhang, D., Wang, A., Sun, R., Luo, X.: A gnn-guided predict-and-search framework for mixed-integer linear programming. arXiv preprint arXiv:2302.05636  (2023)

\bibitem{he2014learning}
He, H., Daume~III, H., Eisner, J.M.: Learning to search in branch and bound algorithms. Advances in neural information processing systems  \textbf{27} (2014)

\bibitem{hendel2022adaptive}
Hendel, G.: Adaptive large neighborhood search for mixed integer programming. Mathematical Programming Computation  \textbf{14},  185--221 (2022)

\bibitem{hosny2024automatic}
Hosny, A., Reda, S.: Automatic milp solver configuration by learning problem similarities. Annals of Operations Research  \textbf{339}(1),  909--936 (2024)

\bibitem{huang2020enhancing}
Huang, T., Dilkina, B.: Enhancing seismic resilience of water pipe networks. In: Proceedings of the 3rd ACM SIGCAS Conference on Computing and Sustainable Societies. pp. 44--52 (2020)

\bibitem{huang2023searching}
Huang, T., Ferber, A.M., Tian, Y., Dilkina, B., Steiner, B.: Searching large neighborhoods for integer linear programs with contrastive learning. In: International Conference on Machine Learning. pp. 13869--13890. PMLR (2023)

\bibitem{huang2024contrastive}
Huang, T., Ferber, A.M., Zharmagambetov, A., Tian, Y., Dilkina, B.: Contrastive predict-and-search for mixed integer linear programs. In: International Conference on Machine Learning. PMLR (2024)

\bibitem{huang2024distributional}
Huang, W., Huang, T., Ferber, A.M., Dilkina, B.: Distributional {MIPLIB}: a multi-domain library for advancing ml-guided milp methods. arXiv preprint arXiv:2406.06954  (2024)

\bibitem{kadioglu2010isac}
Kadioglu, S., Malitsky, Y., Sellmann, M., Tierney, K.: Isac--instance-specific algorithm configuration. In: ECAI 2010, pp. 751--756. IOS Press (2010)

\bibitem{khalil2016learning}
Khalil, E., Le~Bodic, P., Song, L., Nemhauser, G., Dilkina, B.: Learning to branch in mixed integer programming. In: Proceedings of the AAAI Conference on Artificial Intelligence. vol.~30 (2016)

\bibitem{khalil2017learning}
Khalil, E.B., Dilkina, B., Nemhauser, G.L., Ahmed, S., Shao, Y.: Learning to run heuristics in tree search. In: Ijcai. pp. 659--666 (2017)

\bibitem{khalil2022mip}
Khalil, E.B., Morris, C., Lodi, A.: Mip-gnn: A data-driven framework for guiding combinatorial solvers. In: Proceedings of the AAAI Conference on Artificial Intelligence. vol.~36, pp. 10219--10227 (2022)

\bibitem{khalil2022finding}
Khalil, E.B., Vaezipoor, P., Dilkina, B.: Finding backdoors to integer programs: a monte carlo tree search framework. In: Proceedings of the AAAI Conference on Artificial Intelligence. vol.~36, pp. 3786--3795 (2022)

\bibitem{kingma2014adam}
Kingma, D.P., Ba, J.: Adam: A method for stochastic optimization. arXiv preprint arXiv:1412.6980  (2014)

\bibitem{labassi2022learning}
Labassi, A.G., Ch{\'e}telat, D., Lodi, A.: Learning to compare nodes in branch and bound with graph neural networks. Advances in Neural Information Processing Systems  \textbf{35},  32000--32010 (2022)

\bibitem{land2010automatic}
Land, A.H., Doig, A.G.: An automatic method for solving discrete programming problems. Springer (2010)

\bibitem{leyton2000towards}
Leyton-Brown, K., Pearson, M., Shoham, Y.: Towards a universal test suite for combinatorial auction algorithms. In: Proceedings of the 2nd ACM conference on Electronic commerce. pp. 66--76 (2000)

\bibitem{li2024towards}
Li, S., Kulkarni, J., Menache, I., Wu, C., Li, B.: Towards foundation models for mixed integer linear programming. arXiv preprint arXiv:2410.08288  (2024)

\bibitem{lin2024cambranch}
Lin, J., Xu, M., Xiong, Z., Wang, H.: Cambranch: Contrastive learning with augmented milps for branching. arXiv preprint arXiv:2402.03647  (2024)

\bibitem{lindauer2022smac3}
Lindauer, M., Eggensperger, K., Feurer, M., Biedenkapp, A., Deng, D., Benjamins, C., Ruhkopf, T., Sass, R., Hutter, F.: Smac3: A versatile bayesian optimization package for hyperparameter optimization. Journal of Machine Learning Research  \textbf{23}(54), ~1--9 (2022)

\bibitem{liu2019end}
Liu, S., Johns, E., Davison, A.J.: End-to-end multi-task learning with attention. In: Proceedings of the IEEE/CVF conference on computer vision and pattern recognition. pp. 1871--1880 (2019)

\bibitem{lodi2017learning}
Lodi, A., Zarpellon, G.: On learning and branching: a survey. Top  \textbf{25},  207--236 (2017)

\bibitem{nair2020solving}
Nair, V., Bartunov, S., Gimeno, F., Von~Glehn, I., Lichocki, P., Lobov, I., O'Donoghue, B., Sonnerat, N., Tjandraatmadja, C., Wang, P., et~al.: Solving mixed integer programs using neural networks. arXiv preprint arXiv:2012.13349  (2020)

\bibitem{oord2018representation}
Oord, A.v.d., Li, Y., Vinyals, O.: Representation learning with contrastive predictive coding. arXiv preprint arXiv:1807.03748  (2018)

\bibitem{paulus2022learning}
Paulus, M.B., Zarpellon, G., Krause, A., Charlin, L., Maddison, C.: Learning to cut by looking ahead: Cutting plane selection via imitation learning. In: International conference on machine learning. pp. 17584--17600. PMLR (2022)

\bibitem{pohl1970heuristic}
Pohl, I.: Heuristic search viewed as path finding in a graph. Artificial intelligence  \textbf{1}(3-4),  193--204 (1970)

\bibitem{scavuzzo2024machine}
Scavuzzo, L., Aardal, K., Lodi, A., Yorke-Smith, N.: Machine learning augmented branch and bound for mixed integer linear programming. Mathematical Programming pp. 1--44 (2024)

\bibitem{song2020general}
Song, J., Lanka, R., Yue, Y., Dilkina, B.: A general large neighborhood search framework for solving integer programs. In: Annual Conference on Neural Information Processing Systems (NeurIPS) (2020)

\bibitem{song2018learning}
Song, J., Lanka, R., Zhao, A., Bhatnagar, A., Yue, Y., Ono, M.: Learning to search via retrospective imitation. arXiv preprint arXiv:1804.00846  (2018)

\bibitem{tang2020reinforcement}
Tang, Y., Agrawal, S., Faenza, Y.: Reinforcement learning for integer programming: Learning to cut. In: International conference on machine learning. pp. 9367--9376. PMLR (2020)

\bibitem{tarjan1977finding}
Tarjan, R.E., Trojanowski, A.E.: Finding a maximum independent set. SIAM Journal on Computing  \textbf{6},  537--546 (1977)

\bibitem{tong2024optimization}
Tong, J., Cai, J., Serra, T.: Optimization over trained neural networks: Taking a relaxing walk. In: International Conference on the Integration of Constraint Programming, Artificial Intelligence, and Operations Research. pp. 221--233. Springer (2024)

\bibitem{valentin2022instance}
Valentin, R., Ferrari, C., Scheurer, J., Amrollahi, A., Wendler, C., Paulus, M.B.: Instance-wise algorithm configuration with graph neural networks. arXiv preprint arXiv:2202.04910  (2022)

\bibitem{williams2003backdoors}
Williams, R., Gomes, C.P., Selman, B.: Backdoors to typical case complexity. In: IJCAI. vol.~3, pp. 1173--1178 (2003)

\bibitem{xu2011hydra}
Xu, L., Hutter, F., Hoos, H.H., Leyton-Brown, K.: Hydra-mip: Automated algorithm configuration and selection for mixed integer programming. In: RCRA workshop on experimental evaluation of algorithms for solving problems with combinatorial explosion at the international joint conference on artificial intelligence (IJCAI). pp. 16--30 (2011)

\end{thebibliography}

\end{document}